\documentclass[times, review, 10pt]{elsarticle}

\usepackage{amssymb}
\usepackage{amsmath}
\usepackage{multirow}
\usepackage[table,xcdraw]{xcolor}
\usepackage{booktabs}
\usepackage{graphicx}
\usepackage{subcaption}
\usepackage{hyperref}

\journal{Pattern Recognition}

\begin{document}

\begin{frontmatter}



\title{Design Choices for Multimodal Transformer Learning with Missing and Longitudinal Healthcare Data}


\author[1]{Rita Cordeiro Mendes}
\ead{rita.cordeiro@tecnico.ulisboa.pt}
\affiliation[1]{organization={Institute for Systems and Robotics, Instituto Superior Técnico},
            %addressline={}, 
            city={Lisbon},
%          citysep={}, % Uncomment if no comma needed between city and postcode
            %postcode={}, 
            %state={},
            country={Portugal}}

\author[2]{Maria Manuel Pereira}%[]
\affiliation[2]{organization={ULS},
            %addressline={}, 
            city={Braga},
%          citysep={}, % Uncomment if no comma needed between city and postcode
            country={Portugal}}

\author[1]{Carlos Santiago}
\ead{carlos.santiago@tecnico.ulisboa.pt}

\author[1]{Catarina Barata}
\ead{ana.c.fidalgo.barata@tecnico.ulisboa.pt}
            
%% Abstract
\begin{abstract}
Multimodal learning offers the potential to improve predictions by integrating complementary information from heterogeneous data sources. However, multimodal datasets are frequently incomplete, raising important questions about how models should handle missing information. Although recent approaches increasingly rely on Transformer-based architectures for multimodal fusion, incomplete multimodal data introduces challenges such as missing modalities and intra-modal missingness, which motivate key architectural design choices, such as tokenization granularity. In addition, many datasets are longitudinal, with observations collected across multiple time points. The sequential nature of the Transformer architectures provides a natural framework for modeling such data, but introduces additional considerations, such as how temporal information should be represented.
In this study, we systematically investigate these design choices using three different cancer datasets. We compare modality-level and feature-level tokenization and evaluate attention masking and imputation as strategies for handling missing information, both naturally occurring in the datasets and synthetically induced. We further investigate approaches for incorporating temporal information in our training process and how to model it. Experiments are conducted with unimodal and multimodal configurations, as well as different levels of missingness for binary survival classification and multiple time encoding strategies for discrete-time survival prediction. Our results show that while multimodal outperforms unimodal results, missing data heavily impacts predictions and thus effective design choices such as tokenization granularity, and intra-missing handling is essential for reliable prediction.
\end{abstract}


\begin{highlights}
\item Transformer design shape affect multimodal fusion under substantial missingness.
\item Ignoring intra-modal missingness can lead to misleading performance estimates.
\item Explainability using attention scores interpretation is non-trivial.
\item Transformers can effectively model longitudinal multimodal data.
\end{highlights}

%% Keywords
\begin{keyword}
Multimodal Learning \sep Missing Data \sep Transformers


\end{keyword}

\end{frontmatter}



\section{Introduction}\label{introduction}

The rapid growth of digital technologies has led to increasingly large volumes of heterogeneous data across diverse domains. Such data may originate from images, video, audio, sensor signals, time-series measurements, graphs, and other structured or unstructured sources, which often differ in representation, temporal or spatial resolution, completeness, and reliability. Effectively integrating these complementary sources to obtain richer representations than those derived from individual sources alone has therefore become an important challenge in machine learning \citep{deep-multimodal-learning}. However, real-world observations are often noisy, incomplete, asynchronous, and misaligned, while differences in structure and representation can hinder the identification of meaningful correspondences across sources \citep{SMIL}.

These challenges are particularly relevant in healthcare, where a patient's condition may be characterized through clinical measurements, medical imaging, genomic information, and lifestyle or behavioral data. These modalities provide complementary perspectives while differing substantially in structure, scale, temporal characteristics, and reliability \citep{robust-ml-healthcare-survey}. Moreover, individual modalities may contain multiple observations, such as medical images acquired at different time points or anatomical regions. Integrating these sources can therefore provide a more comprehensive representation of patient state.

The digitalization of healthcare has enabled the collection of increasingly large and diverse multimodal clinical datasets, creating new opportunities for deep learning models that support clinical decision-making \citep{multimodal-healthcare}. At the same time, clinical data are frequently incomplete and inherently longitudinal. Medical datasets may contain missing information within modalities or entire modalities due to unperformed examinations, inconsistent data collection, or other practical constraints. Since diseases evolve over time, effective clinical modeling must therefore capture relationships across modalities while accounting for both missing information and temporal structure.

Transformer-based architectures provide a flexible framework for these challenges by representing heterogeneous inputs as a common sequence of tokens and enabling interactions across sources through self-attention. Observations from multiple modalities and time points can thus be integrated within a shared sequence, with positional or temporal encodings capturing their ordering and timing. However, this flexibility introduces important design choices. Missing information must be explicitly handled, for example through masking or imputation, while the choice of tokenization granularity determines how information and missingness are represented. For longitudinal data, the incorporation of temporal information is an additional consideration, whether through dedicated temporal or positional encodings or as part of the input representation.

In this work, we systematically investigate these design choices for Transformer-based multimodal learning under severe missing and incomplete clinical data using three heterogeneous cancer datasets.

The contributions are as follows:
\begin{itemize}
\item We evaluate uni-modal and multimodal configurations across different modality combinations and missingness levels, analyzing when multimodal fusion remains beneficial under incomplete data for both binary survival classification and discrete-time hazard prediction.
\item We investigate the effect of token granularity on multimodal representation and model performance, comparing modality-level and feature-level tokenization.
\item We evaluate attention masking and imputation strategies for handling missingness under both naturally occurring and synthetically induced missingness.
\item We investigate these design choices in longitudinal multimodal data, examining how different strategies for incorporating temporal information affect prediction, particularly when observations occur at irregular time points.
\end{itemize}

\section{Related Work}

Research on multimodal learning has explored a range of strategies for combining information from heterogeneous modalities. Existing approaches primarily differ in how modality-specific representations are constructed and at which stage they are combined \citep{survey-of-multimodal-learning}. More recent approaches have increasingly focused on learning such cross-modal interactions rather than relying solely on concatenation or prediction-level aggregation. Attention mechanisms are particularly suitable for this purpose, as they allow models to learn interactions between elements from different modalities \citep{attention-please} which has motivated the adoption of Transformer-based architectures \citep{multimodal-learning-w-transformers}.

Transformers were originally introduced for sequence modeling \cite{vaswani2023attentionneed} and have since been applied to a wide range of domains \citep{transformers-SOTA-NLP}. For example, the Vision Transformer (ViT) \citep{vit} demonstrated their applicability to image data, while subsequent work has extended Transformers to other domains \citep{transformers-computer-vision}. In multimodal learning, Transformers provide a flexible framework for integrating modality-specific representations and modeling cross-modal interactions, using modality-specific encoders followed by shared Transformer layers or through explicit cross-attention mechanisms \citep{multimodal-learning-w-transformers}.

A key design choice when applying Transformers to heterogeneous data is the definition of the input token. Tokenization determines the granularity at which information is represented and can affect both model performance and computational cost. While NLP models may use words, subwords, or characters \citep{tokenization-LLMS}, other domains require different strategies. ViT, for example, represents images as sequences of patches \citep{vit}, while \citep{effect-tokenization-transformers-bio-sequences} show that tokenization choices affect performance on biological sequences. For structured data, \citep{gorishniy2023revisitingdeeplearningmodels} use feature-level tokenization, representing individual features as tokens. Thus, tokenization is an architectural choice that depends on the structure of the underlying data.

An additional challenge in multimodal learning is incomplete modality availability. Modalities may be missing because of unavailable measurements, sensor failures, or incomplete data collection. A straightforward strategy is to remove incomplete samples \citep{multimodal-missing-remove}, but this can substantially reduce the available training data. Other approaches train directly on incomplete observations, with studies showing that exposure to missing modalities during training can improve robustness at inference time. SMIL \citep{SMIL}, for example, explicitly addresses severely incomplete multimodal data and demonstrates the benefits of learning from incomplete modality configurations. Alternative strategies include imputation \citep{m3care}, reconstructing missing modalities \citep{modality-reconstruction}, or using modality dropout during training \citep{multimodal-dropout}. However, reconstruction becomes increasingly difficult with greater missingness, while modality dropout assumes complete training data.

While many of the aforementioned strategies for handling missing modalities can be incorporated into Transformer-based models, the self-attention mechanism provides an additional strategy for handling missing information: attention masking. In \cite{are-transformers-robust-to-missing-data}, the authors investigate Transformer-based fusion of text and image modalities across three multimodal datasets under varying degrees of modality missingness. They find that training with missing modalities improves performance under both complete and incomplete conditions. However, their analysis considers only modality-level missingness, where an entire modality is unavailable, and does not address missing information within an observed modality.

For longitudinal modeling, Transformer-based architectures have also shown considerable potential for representing the complex temporal structure of clinical data. Several recent approaches have explored the use of Transformers to integrate heterogeneous medical events into unified, time-aware representations of patient trajectories.

Transformers have also been increasingly applied to longitudinal clinical data, where they can integrate heterogeneous medical events into time-aware patient representations. TAMME (Time-Aware MultiModal Transformer Encoder) \citep{TAMME} represents imaging, text, numerical, and categorical information as timestamped events, incorporating the time difference between consecutive events to support irregularly sampled sequences. It also introduces sequence summarization to handle long patient histories. APOLLO \citep{multimodal-temporal-foundation-model} extends this paradigm to large-scale longitudinal patient records by integrating structured events with text and imaging and incorporating learned temporal representations based on patient age. More recently, NOAH \citep{NOAH} uses timestamped multimodal events within a generative Transformer to model and forecast longitudinal patient trajectories.

Together, these approaches demonstrate the potential of representing a patient's longitudinal medical record as a sequence of multimodal, time-aware events. By explicitly incorporating temporal information, they avoid the need for uniformly sampled input sequences and, consequently, for explicit handling of missing events. However, they primarily address missingness at the event or modality level, whereas clinical observations may themselves be only partially observed. Our work therefore explicitly considers intra-modality missingness, allowing partially observed patient records to be represented without treating an entire modality as absent. We investigate how this missingness interacts with multimodal fusion and examine the effects of tokenization and temporal positional encoding on predictive performance and model explainability.

\section{Methodology}

\subsection{Pipeline Overview}
\begin{figure}
    \centering
    \includegraphics[width=0.9\linewidth]{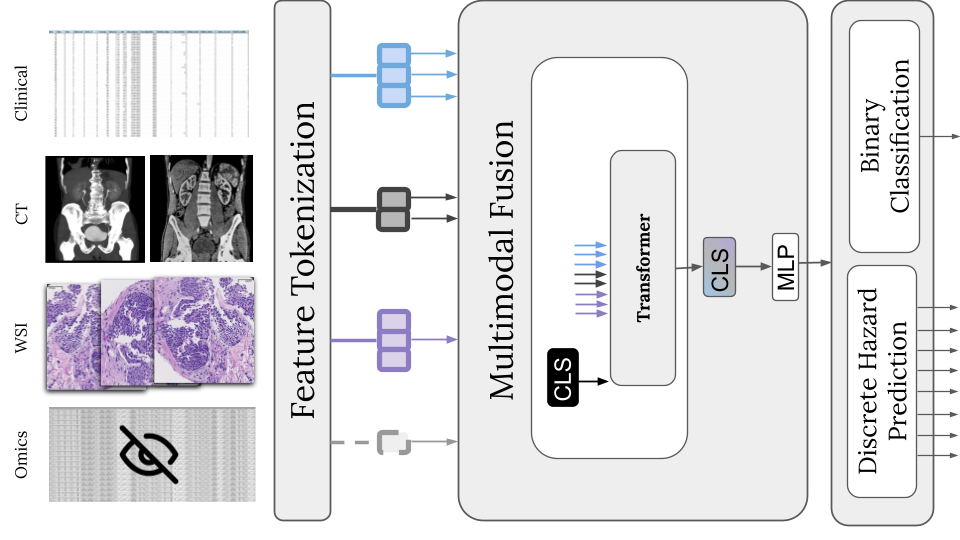}
    \caption{Pipeline overview. Multiple input modalities are represented as an example, showcasing their heterogeneous nature. Omics is represented as missing.}
    \label{pipeline-overview}
\end{figure}

Figure~\ref{pipeline-overview} provides an overview of the proposed architecture. Given a set of heterogeneous input modalities, each modality is first transformed into a sequence of tokens, after which a learnable class token (CLS) is prepended. All tokens are projected into a shared embedding space and provided to a Transformer encoder which enables interactions between tokens, aggregating information into the CLS token. Missing modalities are masked to prevent them from contributing to the representation.

The resulting CLS representation is passed to a prediction head consisting of a linear MLP layer. For classification tasks, the prediction head produces a binary output, whereas for longitudinal analysis, it produces a fixed number of outputs corresponding to predefined temporal bins. The same architecture is used across tasks, with only the output dimensionality of the prediction head differing according to the prediction objective. However, models are trained independently for each task.

\subsection{Transformer-Based Modeling}

The Transformer architecture provides several design choices that are particularly relevant for incomplete multimodal data. First, each modality can be represented by a single token or decomposed into multiple tokens, controlling the granularity at which intra- and inter-modal interactions are modeled. We therefore evaluate different tokenization schemes for multimodal representation.

Second, attention masking allows unavailable inputs to be excluded from the attention computation. We investigate different strategies for handling missing modalities and compare their performance under varying levels of missingness.

Finally, for longitudinal data, we examine how temporal information is incorporated through different positional encoding strategies, comparing learnable and sinusoidal encodings. We further compare multimodal models with their unimodal counterparts and analyze attention scores to investigate how the model utilizes available modalities and to provide insight into its predictions.

\subsection{Multimodal Tokenization}

Before multimodal observations can be processed by a Transformer, they must be transformed into a sequence of tokens with a common embedding dimensionality. The choice of tokenization determines the granularity at which information is represented and therefore influences the interactions that can be learned through self-attention.

\subsubsection{Individual Unit-Level Tokenization}

In the unit-level approach, the basic unit represented by an individual token depends on the modality type. For tabular data, we can represent each feature within a modality as an individual token, whereas for imaging, each scan can be an individual token. Let $X^{(m)}$ represent the features for modality $m$, where $D_m$ is the number of features or scans. Each modality is then transformed into a sequence of $D_m$ tokens:
\begin{equation}
    X^{(m)}
    \rightarrow
    [\mathbf{z}^{(m)}_1,
    \mathbf{z}^{(m)}_2,
    \ldots,
    \mathbf{z}^{(m)}_{D_m}].
\end{equation}

Each feature is projected into a shared embedding space of dimensionality $d$, and a learned feature-specific embedding is added to each token, allowing the Transformer to model relationships between individual features both within and across modalities. However, representing each feature as a separate token causes the sequence length to grow with the dimensionality of the input, substantially increasing the computational cost of self-attention and potentially making this representation impractical for high-dimensional data.
Despite this computational cost, feature-level tokenization offers several advantages. In particular, it enables individual features to be selectively masked, which is useful for representing missing or unavailable inputs at a fine-grained level. Moreover, treating features as separate tokens facilitates a more detailed analysis of model behavior, as the contribution and interactions of individual features can be examined through the self-attention matrix. Thus, feature-level tokenization provides greater flexibility in handling partially observed inputs and interpretability, at the expense of increased computational complexity.

\subsubsection{Modality-Level Tokenization}

Alternatively, an entire modality can be represented by a single token. In this approach, the complete modality is mapped to a shared embedding space using a learnable modality-specific projection function $f_m$:
\begin{equation}
    \mathbf{z}^{(m)} = f_m(X^{(m)}),
\end{equation}
where $\mathbf{z}^{(m)} \in \mathbb{R}^{d}$ represents the token corresponding to modality $m$. The resulting multimodal input sequence for M modalities is then given by an input sequence of M tokens such as
\begin{equation}
    \mathbf{Z}
    =
    [
    \mathbf{z}^{(1)},
    \mathbf{z}^{(2)},
    \ldots,
    \mathbf{z}^{(M)}
    ].
\end{equation}

This representation substantially reduces the sequence length and treats each modality as a high-level information unit. However, interactions between individual features are modeled within the modality-specific projection function before multimodal attention is applied, potentially limiting fine-grained interactions between features inter and intra modalities. Moreover, this method only allows for modality-level masking, which is fine if a whole modality is missing but lacks when a modality is incomplete.

\subsection{Missing Data Handling}

Attention masking provides a mechanism for selectively excluding specific tokens from the attention computation. By applying a mask to the corresponding entries of the attention matrix, the model is prevented from attending to tokens associated with unavailable information, thereby ensuring that missing observations do not contribute to the resulting representation. In this work, attention masking is used as the primary strategy for handling missing data. When an observation is unavailable, its corresponding token is masked during both training and inference. This allows samples with incomplete observations to be retained rather than discarded, enabling the model to learn from all available information while remaining robust to different patterns of missingness.

As a baseline, we additionally evaluate an imputation-based strategy in which missing observations are replaced with a predefined meaningless value. This approach allows the model to receive a complete input while potentially retaining information about the absence of the original observation. Such an explicit representation of missingness can be informative, especially when it is related to the underlying patient characteristics or clinical process, as may occur under missing-not-at-random mechanisms.

\subsection{Longitudinal Modeling}

Transformers were originally introduced for sequence modeling and are therefore well suited to modeling longitudinal observations represented as ordered sequences \cite{vaswani2023attentionneed}. Initial approaches used sinusoidal positional encodings to provide information about the position of each element within a sequence. In a longitudinal setting, positional encodings allow the model to distinguish observations at different positions and capture dependencies across the sequence. The choice of positional encoding depends on the tokenization strategy and modality structure. For modality-level tokenization of tabular data, observation time can be incorporated directly as an additional feature. However, this is less straightforward when modalities contain multiple tokens or are non-tabular. Positional or temporal encodings can instead be applied to individual tokens, providing a consistent mechanism for representing temporal information across modalities.

The original Transformer uses fixed sinusoidal positional encodings, which assign each sequence position a unique representation based on sinusoidal functions with different wavelengths across the representation dimensions. The formulation is described in \cite{vaswani2023attentionneed}. Although sinusoidal and learnable encodings performed similarly, the authors favoured sinusoidal encoding because it can generalise to sequence lengths beyond those observed during training.

However, positional encoding represents sequence order rather than actual temporal intervals. This is a limitation for clinical data, where observations are often irregularly and sparsely sampled. Consecutive observations may therefore be separated by substantially different time intervals despite having adjacent positional indices. To account for this irregular temporal structure, we replace fixed positional encoding with a learnable temporal encoding based on the elapsed time of each observation. Given an elapsed time $t$, we first apply a logarithmic transformation to compress large intervals while preserving differences between observations close in time. The transformed value is then mapped to a $d_{model}$-dimensional representation using a learnable multilayer perceptron (MLP). Thus, the mapping from time to representation is learned directly from the data rather than predetermined by fixed frequencies.

For an event occurring at time $t$, the token representation is therefore given by
\begin{equation}
\begin{aligned}
    z=e_{\text{feature}}+e_{\text{modality}}+\underbrace{\text{MLP}(\log(1+t))}_{\text{Learnable PE(t)}}
\end{aligned}    
\end{equation}

\section{Experimental Setup}

This section describes the datasets, training and evaluation protocols, and an overview of experiments used to systematically assess best practices when designing Transformer-based models for multimodal data with substantial missingness. The evaluation considers three cancer types: Lung, Kidney, and Gastric cancer. The datasets differ in patient populations, modality availability, and missingness patterns, enabling evaluation across heterogeneous clinical settings.

\subsection{Lung Cancer Multimodal Dataset}
The MMIST-Lung dataset \cite{mmist-Lung} comprises 1,365 patients from the CPTAC and TCGA cohorts with Lung adenocarcinoma (LUAD) \citep{TCGA-LUAD} \citep{CPTAC-LUAD} and Lung squamous cell carcinoma (LUSC) \citep{TCGA-LUSC} \citep{CPTAC-LSCC}. It integrates clinical information (CLIN), transcriptomics (TRNSC), follow-up data (FUP), treatment information including chemotherapy (CHEM), radiotherapy (RAD), surgery (SURG), and immunotherapy (IMMUNO), as well as imaging data, including whole-slide imaging (WSI), computed tomography (CT), and positron emission tomography (PET). For the imaging modalities, WSI features were extracted using TITAN \citep{TITAN}, while CT and PET features were obtained using MedNet \citep{MedNet}. Patient and modality-level availability are summarised in Table~\ref{tab:Lung-distribution}. 

CLIN is available for all patients and includes demographic and tumor-related characteristics, including age, gender, race, and AJCC \citep{ajcc} staging, as well as smoking-related information. Although CLIN contains missing values at the feature level, it is used as the unimodal baseline throughout the experiments because it represents the most consistently available and informative modality.

The dataset also contains longitudinal information. CLIN, TRNSC, and imaging modalities are assumed to correspond to the time of diagnosis, while follow-up and treatment data are associated with subsequent time points represented by days elapsed since diagnosis. Of the 1{,}365 patients, 671 have a recorded last known contact date and are therefore available for longitudinal hazard prediction.

Patients may have multiple scans within each imaging modality. WSI is available for nearly all patients and comprises diagnostic (DX), top (TS), and bottom (BS) slides, with a median of three slides per patient. CT and PET are considerably sparser, with at least one scan available for 71 and 33 patients, respectively, and median scan counts of five and four per patient.

\begin{table}[]
\centering
\caption{Lung cancer modality patient distribution, comprising clinical (CLIN), transcriptomics (TRNSC), follow-up (FUP), treatment data (CHEM, RAD, SURG, IMMUNO) and imaging data (WSI, CT, PET).}
\label{tab:Lung-distribution}
\fontsize{8pt}{9.6pt}\selectfont
\begin{tabular}{@{}ccc@{}}
\toprule
\cellcolor[HTML]{FFFFFF}Modality &
  \cellcolor[HTML]{FFFFFF}\begin{tabular}[c]{@{}c@{}}Patient\\ Number\end{tabular} &
  \begin{tabular}[c]{@{}c@{}}Modality\\ Missingness\end{tabular} \\ \midrule
CLIN   & 1365 & 0     \\ 
TRNSC  & 1284 & 6\%   \\ 
FUP    & 880  & 35\%  \\ 
CHEM   & 277  & 80\%  \\
RAD    & 923  & 32\%  \\
SURG   & 291  & 79\%  \\
IMMUNO & 49   & 96\%  \\ 
WSI    & 1359 & 0.4\% \\
CT     & 71   & 95\%  \\
PET    & 33   & 98\%  \\ \bottomrule
\end{tabular}
\end{table}

\subsection{Kidney Cancer Multimodal Dataset}
The MMIST-ccRCC dataset \citep{mmist-ccrcc} comprises 618 patients with clear cell renal cell carcinoma from the CPTAC-CCRCC \citep{CPTAC-CCRCC} and TCGA-KIRC \citep{TCGA-KIRC} cohorts. It includes clinical information (CLINGEN, referred to as CLIN in this work), 10{,}270 selected transcriptomic genes (TRNSC) and three imaging modalities: WSI, CT, and MRI. Modality availability and distribution is reported in Table~\ref{tab:Kidney-distribution}. Clinical and WSI data are available for all patients, while TRNSC is available for the majority of the cohort. Imaging modalities are considerably sparser, particularly MRI. When available, imaging modalities may contain multiple scans per patient.

CLIN contains 11 demographic and tumour-related features and is used as the unimodal baseline due to its complete availability. The dataset contains no explicit temporal information and therefore supports only binary prediction of one-year overall survival. The target is highly imbalanced, with 88\% of patients surviving to one year.

\begin{table}[]
\centering
\caption{Modality distribution for the Kidney cancer cohort, consisting of clinical (CLIN), selected transcriptomics (TRNSC), WSI, CT and MRI data.}
\label{tab:Kidney-distribution}
\fontsize{8pt}{9.6pt}\selectfont

\begin{tabular}{@{}ccc@{}}
\toprule
\cellcolor[HTML]{FFFFFF}Modality &
  \cellcolor[HTML]{FFFFFF}\begin{tabular}[c]{@{}c@{}}Patient\\ Number\end{tabular} &
  \begin{tabular}[c]{@{}c@{}}Modality\\ Missingness\end{tabular} \\ \midrule
CLIN & 618 & 0    \\
TRNSC   & 614 & 99\% \\
WSI     & 618 & 0    \\
CT      & 256 & 59\% \\
MRI     & 56  & 91\% \\ \bottomrule
\end{tabular}
\end{table}

\subsection{Gastric Cancer Multimodal Dataset}
The Gastric cancer dataset is a private, in-house multimodal dataset comprising 274 patients aged 18 years or older who were diagnosed with gastric cancer and treated between January 2019 and October 2024. This study was approved by the Ethics Committee of Unidade Local de Saúde de Braga, EPE (approval no. 158\_2025) and conducted in accordance with the Declaration of Helsinki. Given its retrospective design and use of anonymized data, informed consent was waived. 

The distribution of patients across modalities is presented in Table~\ref{tab:Gastric-distribution}. Unlike the Lung and Kidney cancer datasets, this is a strictly tabular dataset and does not contain imaging data. Demographic (DEM) data include patient characteristics, symptoms, risk factors, and comorbidities while neoplasia (NEO) data contain tumor-related information, including AJCC TNM staging, tumor location, and Lauren classification \citep{lauren}. A subset of patients developed metastatic (MTX) disease, either at the time of diagnosis or during follow-up. Demographic and neoplasia information are available for all patients.

Treatment data include palliative treatment (PAL), up to three lines of therapy, and curative treatments including surgery (SURG), adjuvant chemotherapy (ADJ CHEM), perioperative chemotherapy (PERIOP CHEM), and other treatments (OTHER TX). All clinical and treatment events are associated with precise dates, enabling patient-specific longitudinal timelines and analysis of disease progression and treatment patterns. Entries marked as "Not Applicable" are treated as missing data, despite potentially containing informative information.

\begin{table}[]
\centering
\caption{Modality distribution for the Gastric cancer cohort.}
\label{tab:Gastric-distribution}
\fontsize{8pt}{9.6pt}\selectfont
\begin{tabular}{ccc}
\hline
{\color[HTML]{000000} Modality} &
  {\color[HTML]{000000} \begin{tabular}[c]{@{}c@{}}Patient\\ Number\end{tabular}} &
  {\color[HTML]{000000} \begin{tabular}[c]{@{}c@{}}Modality\\ Missingness\end{tabular}} \\ \hline
{\color[HTML]{000000} DEM} & {\color[HTML]{000000} 274} & {\color[HTML]{000000} 0} \\
{\color[HTML]{000000} NEO} & {\color[HTML]{000000} 274} & {\color[HTML]{000000} 0} \\
MTX                        & 161                        & 59\%                     \\
PAL                        & 141                        & 51\%                     \\
ADJ CHEM                  & 46                         & 17\%                     \\
PERIOP CHEM               & 116                        & 42\%                     \\
SURG                       & 162                        & 59\%                     \\
EXTRA TX                  & 274                        & 0                        \\ \hline
\end{tabular}
\end{table}

\subsection{Training and Evaluation Protocol}
\textbf{Hyperparameters and model architecture.} The model configuration and hyperparameters were selected using OPTUNA \citep{optuna}, with validation-set balanced accuracy as the optimization objective. The search space included the learning rate and the main architectural and regularization parameters such as embedding dimension and number of Transformer heads and layers. Empirically, the learning rate was found to be the most sensitive hyperparameter, having the greatest impact on performance. Therefore, proper tuning is essential for stable convergence and generalization. For the Lung cohort, we used a learning rate of $lr=1.3\times10^{-4}$ and $lr=6.98\times10^{-4}$ for the Kidney and Gastric cohorts.

The proposed architecture uses modality-specific projections followed by Transformer\allowbreak-based fusion. All modalities are projected into a shared embedding space with $d_{model}=128$ for Kidney and $d_{model}=256$ for Lung and Gastric. The Transformer encoder is consistent across datasets and consists of two sequential layers, each comprising a four-head multi-head self-attention layer and a position-wise feed-forward network. Following self-attention, dropout ($p=0.2$), a residual connection, and layer normalisation are applied. The representation is then passed through a feed-forward network with dimensions $d_{model}\rightarrow2048\rightarrow d_{model}$, followed by the same dropout, residual connection, and layer normalisation. The two encoder layers are applied sequentially, with the output of the first serving as the input to the second. This relatively compact architecture was chosen to reduce overfitting given the limited dataset sizes while retaining sufficient capacity for multimodal learning. The model is trained using the Adam optimiser. For binary classification, we use Binary Cross Entropy loss with class weights to address class imbalance. For discrete-time survival analysis, we use a discrete-time survival negative log-likelihood based on the predicted hazard for each time interval. For patients experiencing the event, the loss accounts for survival through all preceding intervals and the probability of experiencing the event in the observed interval. For censored patients, it accounts only for survival through the censoring interval, without assuming that the event occurs thereafter.

\textbf{Evaluation Metrics.} The classification experiments were evaluated using Balanced Accuracy (BAcc) and Area Under the Receiver Operating Characteristic Curve (ROC-AUC), with strong performance reflected by high values for both metrics. For the longitudinal analysis, we used the Concordance Index (C-Index) and Brier Score, which provide complementary measures of survival prediction. The C-Index evaluates the model's ability to discriminate between patients with different survival times, with higher values indicating better discrimination. The Brier Score measures the accuracy of predicted survival probabilities by comparing predicted and observed outcomes over time, with lower values indicating better accuracy. Due to the limited number of samples, stratified 5-fold cross-validation was used to ensure stable training. Balanced accuracy on the validation set was monitored for early stopping. Stratification was performed based on the prediction task only and did not account for the distribution of modalities across folds.

\textbf{Explainability.} The Transformer's attention mechanism provides a means of examining how the model distributes attention across input tokens. Since each layer and attention head produces its own attention scores, these can be analysed at different levels of granularity, such as individual heads or aggregated across heads and layers. However, attention scores alone may not reliably reflect the contribution of individual tokens to the final prediction, as representations become increasingly mixed across successive layers. To address this limitation, \cite{rollout} propose attention rollout, which recursively combines attention weights across layers to account for the propagation of information through the network. We therefore use attention rollout as the primary basis for analysing model attention and interpreting its behaviour.

\subsection{Experiments Overview}

Using the three cancer datasets described above, we evaluate Transformer-based multimodal models across four main questions: (i) whether multimodal fusion improves over unimodal models; (ii) which tokenization strategy best represents heterogeneous modalities; (iii) how missing-modality handling affects performance and robustness; and (iv) how longitudinal information can be incorporated to improve multimodal predictions.

\textbf{Classification:} We first perform binary survival classification on the Lung and Kidney cohorts using only information available at diagnosis. For Lung, the target is disease-specific survival, while for Kidney, it is vital status at one year. We then repeat the classification task on the Lung and Gastric cohorts using the complete longitudinal patient sequence to incorporate all available events and measurements.

\textbf{Survival Analysis:} We additionally perform discrete-time longitudinal survival analysis on the Lung and Gastric cohorts. Unlike binary classification, this task models survival over time while accounting for right-censored observations. Survival is predicted over a five-year horizon discretized into 260 weekly intervals. Two landmarking strategies are considered. First, predictions are made relative to each patient's last known medical event, using all information available up to that point. Second, a fixed one-year landmark is used, including only patients who remain event-free at that time, to provide a common temporal reference point. For both settings, we compare fixed sinusoidal positional encoding with the proposed learnable temporal positional encoding.

Together, these experiments evaluate multimodal fusion, tokenization, missing-modality handling, and temporal modelling across datasets with different modality compositions and longitudinal structures.

\section{Results}

\subsection{Unimodality vs. Multimodality}

We first evaluate whether multimodal fusion improves performance over a unimodal baseline under substantial missingness. To isolate the effect of fusion, each modality is represented by a single token, with missing features imputed using a placeholder value before tokenisation. No additional missing-data strategy is applied. Results are shown in the left side of Table~\ref{tab:uni-modal}, under Single-Token. For both Lung and Kidney, the unimodal baseline uses clinical information available at diagnosis, while multimodal configurations combine clinical, imaging, and transcriptomic data.

Multimodal models provide only limited and inconsistent improvements over the clinical baseline and generally exhibit higher standard deviations. This may reflect the strong predictive value and complete availability of the clinical modality, particularly features related to tumour stage and disease severity, leaving limited additional information for the other modalities to contribute.

To assess the importance of these clinical predictors and the effect of feature-level missingness, we repeat the experiment after masking the AJCC staging variables ($T$, $N$, $M$, and pathological stage) which are expected to be among the most informative clinical features for survival prediction. Results are shown on the right side of Table~\ref{tab:uni-modal}. Removing these variables leads to a clear degradation in performance across all configurations and cancer cohorts, confirming both their substantial contribution to the prediction task and the impact of feature-level missingness on model performance. However, multimodal models retain comparatively stronger ROC-AUC performance, increasing the performance gap between unimodal and multimodal configurations, suggesting that when key clinical predictors are unavailable, additional modalities can partially compensate for the loss of clinical information.

It is important to note that this ablation represents a deliberately challenging setting. Although the AJCC variables contain some naturally occurring missing values, they are rarely missing simultaneously; their complete removal therefore represents a stronger form of corruption than the naturally occurring feature-level missingness.

\begin{table}[]
\centering
\caption{Single-token encoding results for unimodal (CLIN) and multimodal configurations incorporating clinical (CLIN), imaging (IMGS), and transcriptomic (TRNSC) data for Lung and Kidney cancer. The right panel shows the ablation of variables related to AJCC tumor information.}
\label{tab:uni-modal}
\fontsize{8pt}{9.6pt}\selectfont
\begin{tabular}{@{}llcccc@{}}
\toprule
                         &                 & \multicolumn{2}{c}{Single-Token} & \multicolumn{2}{c}{AJCC Ablation} \\ \midrule
\cellcolor[HTML]{FFFFFF}\begin{tabular}[c]{@{}l@{}}Cancer\\ Cohort\end{tabular} &
  \cellcolor[HTML]{FFFFFF}\begin{tabular}[c]{@{}l@{}}Modality\\ Configuration\end{tabular} &
  BalAcc {[}\%{]} &
  ROCAUC {[}\%{]} &
  BalAcc {[}\%{]} &
  ROCAUC {[}\%{]} \\ \midrule
                         & CLIN            & 69.45 ± 1.34    & 71.58 ± 1.56   & 50.59 ± 1.53    & 58.19 ± 3.41    \\
\multirow{-2}{*}{Lung}   & CLIN+IMGS+TRNSC & 71.65 ± 5.75    & 75.97 ± 2.98   & 57.38 ± 5.32    & 71.17 ± 3.40    \\
                         & CLIN            & 80.89 ± 3.66    & 86.20 ± 5.35   & 59.89 ± 5.97    & 63.04 ± 8.19    \\
\multirow{-2}{*}{Kidney} & CLIN+IMGS+TRNSC & 78.68 ± 5.51    & 82.55 ± 4.45   & 69.94 ± 6.33    & 74.73 ± 7.20    \\
\bottomrule
\end{tabular}
\end{table}

\subsection{Missing Data Handling: Attention vs Imputation}

The results presented in the previous subsection suggest that missing information can affect multimodal predictions, particularly when highly informative clinical features are unavailable. To explicitly address missing data, we exploit the masking mechanism inherent to Transformer architectures. By masking individual tokens, the model can be prevented from attending to missing features, allowing incomplete observations to be handled without introducing an artificial value into the representation.

To fully leverage this mechanism, we adopt a multi-token representation in which each clinical tabular feature is represented by an individual token. This allows missing features to be selectively masked while retaining the observed features. For imaging data, we likewise adopt a multi-token representation, with each scan represented as an individual token. The specific tokenization strategy used for imaging and its implications are described in the following subsection. However, applying the same strategy to transcriptomic data would be computationally unfeasible given the large number of genes represented in this modality. Importantly, transcriptomic data are complete at the feature level in our experiments and therefore do not introduce feature-level missingness. Consequently, although transcriptomics is included in the multimodal setting, it does not require the same missing-data handling as the other modalities. 

We compare two strategies for handling missing features: attention masking and imputation. With attention masking, tokens corresponding to missing features are explicitly masked and therefore cannot contribute to the Transformer representations. With imputation, missing features are instead replaced with a designated value and remain accessible to the attention mechanism. Consequently, the model can potentially learn from the pattern of missingness itself, in addition to the observed feature values. The results across the Lung and Kidney cancer cohorts are presented in Table~\ref{tab:attn-imp}.

Under naturally occurring missingness, the two strategies produce relatively similar results. In fact, for the Lung cancer cohort, imputation achieves the best performance in both balanced accuracy and area under the ROC curve for the unimodal clinical model. This suggests that the model may be able to exploit the missingness pattern as an additional predictive signal.

\begin{table}[]
\caption{Results for missing data handling attention or imputation strategies for all cancer cohorts and unimodal versus multimodal configurations.}
\label{tab:attn-imp}
\fontsize{8pt}{9.6pt}\selectfont
\begin{tabular}{@{}llcccc@{}}
\toprule
\multirow{2}{*}{\begin{tabular}[c]{@{}l@{}}Cancer\\ Cohort\end{tabular}} & \multirow{2}{*}{\begin{tabular}[c]{@{}l@{}}Modality\\ Configuration\end{tabular}} & \multicolumn{2}{c}{Masking}                   & \multicolumn{2}{c}{Imputation}                \\ \cmidrule(l){3-6} 
                               &                                                                                   & BalAcc {[}\%{]}       & ROCAUC {[}\%{]}       & BalAcc {[}\%{]}       & ROCAUC {[}\%{]}       \\ \midrule
\multirow{2}{*}{Lung}          & CLIN                                                                              & 61.57 ± 1.75          & 65.20 ± 2.86          & \textbf{65.96 ± 2.88} & \textbf{66.92 ± 3.15} \\
                               & CLIN+IMGS+TRNSC                                                               & \textbf{68.26 ± 1.79}          & \textbf{72.15 ± 2.14}          & 66.91 ± 2.75 & 71.64 ± 4.50 \\
\multirow{2}{*}{Kidney}        & CLIN                                                                              & \textbf{81.32 ± 3.82} & \textbf{84.90 ± 4.15} & 79.06 ± 4.38          & 83.97 ± 4.60          \\
                               & CLIN+IMGS+TRNSC                                                               & \textbf{81.15 ± 3.43} & \textbf{85.01 ± 3.91} & 75.22 ± 3.93          & 80.07 ± 3.76          \\
 \bottomrule
\end{tabular}
\end{table}

To further investigate the behavior of the two strategies under more severe missingness, we perform an additional ablation at inference time. Specifically, we deliberately mask all AJCC staging information ($T$, $N$, $M$, and overall pathological staging), removing some of the most informative features related to tumor extent and disease severity. The results are shown in Table~\ref{tab:attn-imp-ablation}. Under this substantially more challenging setting, the difference between the two missing-data strategies becomes more pronounced. This experiment provides a stronger test of whether the model can remain robust when a highly informative subset of clinical features is completely unavailable, rather than merely relying on naturally occurring, relatively sparse feature-level missingness.

Importantly, revisiting the single-token strategy introduced in the previous section, these ablation results highlight how this representation can be misleading. Although the results from Table~\ref{tab:attn-imp} and Table~\ref{tab:uni-modal} are relatively similar, with the single-token approach even achieving slightly better performance for the Lung cohort, this advantage does not persist when evaluating the AJCC ablations. In the Lung cohort, the multimodal masking strategy achieves approximately 66\% balanced accuracy, compared with 57\% for the single-token approach, corresponding to an improvement of around 9 percentage points. The gain is less pronounced in the Kidney cohort, potentially reflecting differences in the underlying patterns of missingness across datasets. Notably, the multi-token strategies also exhibit lower variation across folds, suggesting more consistent performance.

\begin{table}[]
\centering
\caption{Balanced accuracy results, reported as percentages, for the AJCC ablation study comparing masking and imputation strategies for handling missing data. Improvements obtained through masking are highlighted in {\color[HTML]{009901} green}. ALL denotes clinical, imaging and transcriptomics, supressed for spacing.}
\label{tab:attn-imp-ablation}
\fontsize{8pt}{9.6pt}\selectfont
\begin{tabular}{@{}ccccc@{}}
\toprule
\begin{tabular}[c]{@{}c@{}}Cancer\\ Cohort\end{tabular} &
  \begin{tabular}[c]{@{}c@{}}Modality\\ Config.\end{tabular} &
  {\color[HTML]{000000} Masking} &
  Imputation &
  $\Delta$Masking $-$ Imputation \\ \midrule
                         & CLIN & {\color[HTML]{000000} 60.65 ± 1.44} & 57.22 ± 5.27 & {\color[HTML]{009901} 3.43\%} \\
\multirow{-2}{*}{Lung}   & ALL  & 65.86 ± 1.84                        & 62.69 ± 3.61 & {\color[HTML]{009901} 3.17\%} \\
                         & CLIN & 60.15 ± 4.83                        & 58.26 ± 7.99 & {\color[HTML]{009901} 1.89\%} \\
\multirow{-2}{*}{Kidney} & ALL  & 67.03 ± 8.97                        & 67.78 ± 8.30 & -0.75\%                       \\ \bottomrule
\end{tabular}
\end{table}

\begin{figure}
    \centering
    \includegraphics[width=0.9\linewidth]{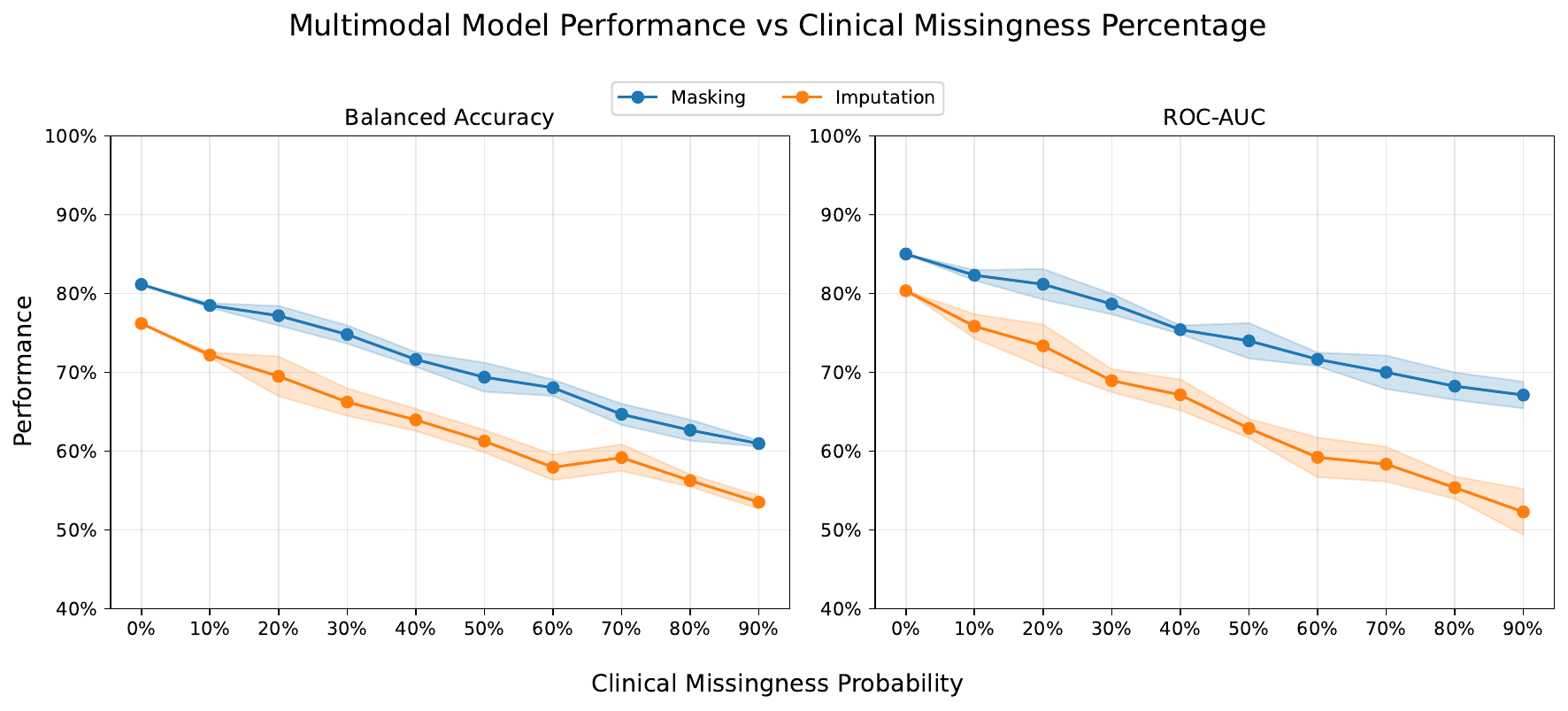}
    \caption{Performance of the multimodal model under increasing levels of randomly introduced clinical feature missingness in the Kidney cohort comparing masking and imputation strategies.}
    \label{fig:Kidney-missingness-percentage}
\end{figure}

So far, these ablation studies considered only specific scenarios in which AJCC tumor information was missing. To further assess the robustness of the missing-data handling strategies, Figure~\ref{fig:Kidney-missingness-percentage} shows the performance of the multimodal model across increasing levels of missing clinical data. In this experiment, features are randomly selected for removal rather than being restricted to the AJCC subset, with the proportion of missing features increasing according to the specified missingness level. To ensure that the results were not driven by the specific features selected for removal, we conducted five runs using different random seeds. This feature rotation exposes the models to different patterns of missingness at each run, allowing us to assess the effect of missingness independently of the importance of any particular set of features. The results show that masking consistently outperforms imputation across the different missingness levels, for both balanced accuracy and ROC-AUC metrics.

\subsection{Image Modalities Tokenization}

While individual features can be represented as separate tokens to handle intra-modality missingness, imaging modalities require a different strategy because the number and type of scans available within the same modality may vary between patients. For example, within the Lung cohort, CT may include chest or abdomen scans, with or without contrast, while PET may include abdomen or whole-body scans. Similarly, Kidney CT may include abdomen, chest, or pelvis scans, with or without contrast. Thus, although an imaging modality may be available for a patient, the specific scans observed within that modality can vary.

We therefore compare two strategies: representing all scans within each modality as a single token using average pooling, or representing each scan as a separate token. Table~\ref{tab:img-tkn} compares these approaches for the Lung and Kidney cohorts. Predictive performance is broadly similar, particularly for multimodal configurations and WSI-only models, with somewhat larger differences when CT, PET, and MRI are included. Despite the comparable performance, we favour representing each scan as a separate token because this preserves scan-level information and provides greater interpretability.

Representing scans individually allows attention rollout to identify the contribution of specific scans, whereas aggregation prevents distinguishing between informative and less informative scans. This is particularly relevant for WSI, where different slide types may provide complementary information. Figure~\ref{Kidney-rollout} illustrates this distinction, showing that the WSI-2 scan receives substantially less attention than the other scans for a particular patient. Such scan-level analysis would not be possible if all scans within a modality were represented by a single aggregated token.

\begin{table}[]
\centering
\caption{Imaging modalities tokenization strategy comparison for Lung and Kidney cancer cohorts.}
\label{tab:img-tkn}
\fontsize{8pt}{9.6pt}\selectfont
\begin{tabular}{@{}llcccc@{}}
\toprule
\multirow{2}{*}{\begin{tabular}[c]{@{}l@{}}Cancer\\ Cohort\end{tabular}} &
\multirow{2}{*}{\begin{tabular}[c]{@{}l@{}}Modality\\ Configuration\end{tabular}} &
\multicolumn{2}{c}{Token per Scan} &
\multicolumn{2}{c}{Token per Modality} \\
\cmidrule(lr){3-4} \cmidrule(l){5-6}
& & BalAcc [\%] & ROCAUC [\%] & BalAcc [\%] & ROCAUC [\%] \\
\midrule
\multirow{2}{*}{Lung}
& WSI
& 69.13 ± 3.14
& 72.35 ± 2.84
& 71.73 ± 3.77
& 74.87 ± 3.64 \\
& CT+PET+WSI
& 69.52 ± 2.19
& 73.32 ± 2.68
& 69.29 ± 2.77
& 71.23 ± 3.76 \\
\midrule
\multirow{2}{*}{Kidney}
& WSI
& 76.86 ± 7.45
& 76.62 ± 10.85
& 76.69 ± 5.50
& 78.37 ± 6.00 \\
& CT+PET+WSI
& 80.60 ± 7.84
& 82.35 ± 8.25
& 76.28 ± 3.95
& 78.56 ± 4.05 \\
\bottomrule
\end{tabular}
\end{table}

\subsection{Attention Rollout Analysis}

\begin{figure}
    \centering
    \includegraphics[width=0.98\linewidth]{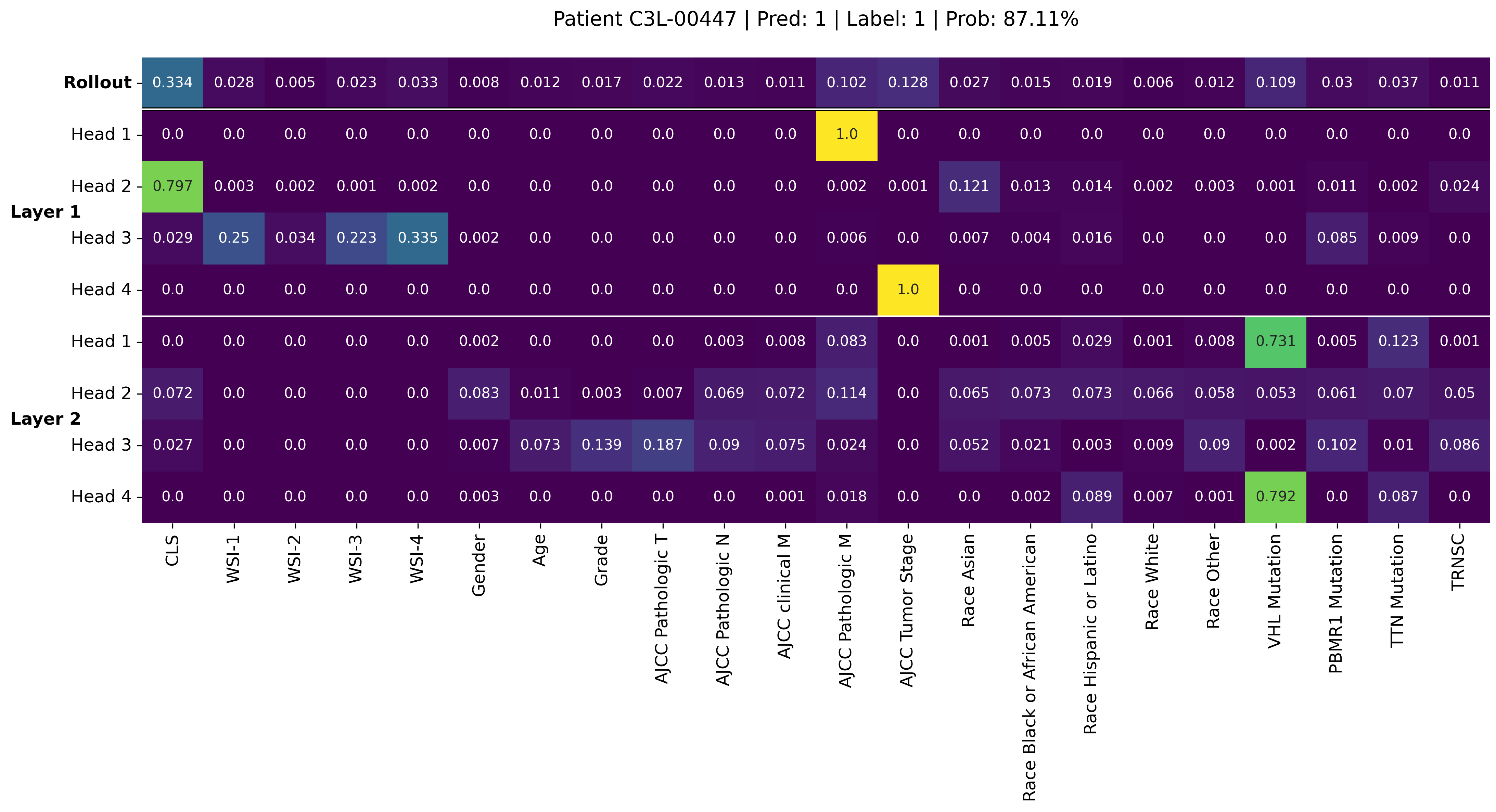}
    \caption{Attention scores analysis for a Kidney cancer cohort patient.}
    \label{Kidney-rollout}
\end{figure}

Figure~\ref{Kidney-rollout} presents attention rollout scores alongside individual attention scores for each head and Transformer layer for a representative Kidney cancer patient. The individual attention maps illustrate how different attention heads attend to different input tokens, highlighting substantial variation in attention patterns across heads and layers. In the adopted tokenization scheme, each WSI scan and clinical feature is represented by a separate token, while the transcriptomic modality is represented by a single token containing all selected genes. Missing features are masked and therefore do not appear in the attention heatmaps.

A common approach to interpreting Transformer attention is to average the attention scores across heads in the final layer. However, this may overlook information propagated through earlier layers. For example, in Figure~\ref{Kidney-rollout}, the final-layer heads would assign zero attention to all WSI tokens which could incorrectly suggest that WSI information does not contribute to the prediction. However, Head 2 of the first layer attends to WSI tokens, indicating that this information is incorporated earlier in the network.

Attention rollout provides a more comprehensive view by recursively aggregating attention across layers. As shown in Figure~\ref{Kidney-rollout}, rollout assigns non-zero attention to the input tokens despite some individual heads assigning zero attention. This demonstrates how information can be propagated across layers and why individual attention maps or final-layer averages may not fully capture the contribution of different modalities.

\begin{figure}[t]
    \centering
    \includegraphics[width=0.9\linewidth]{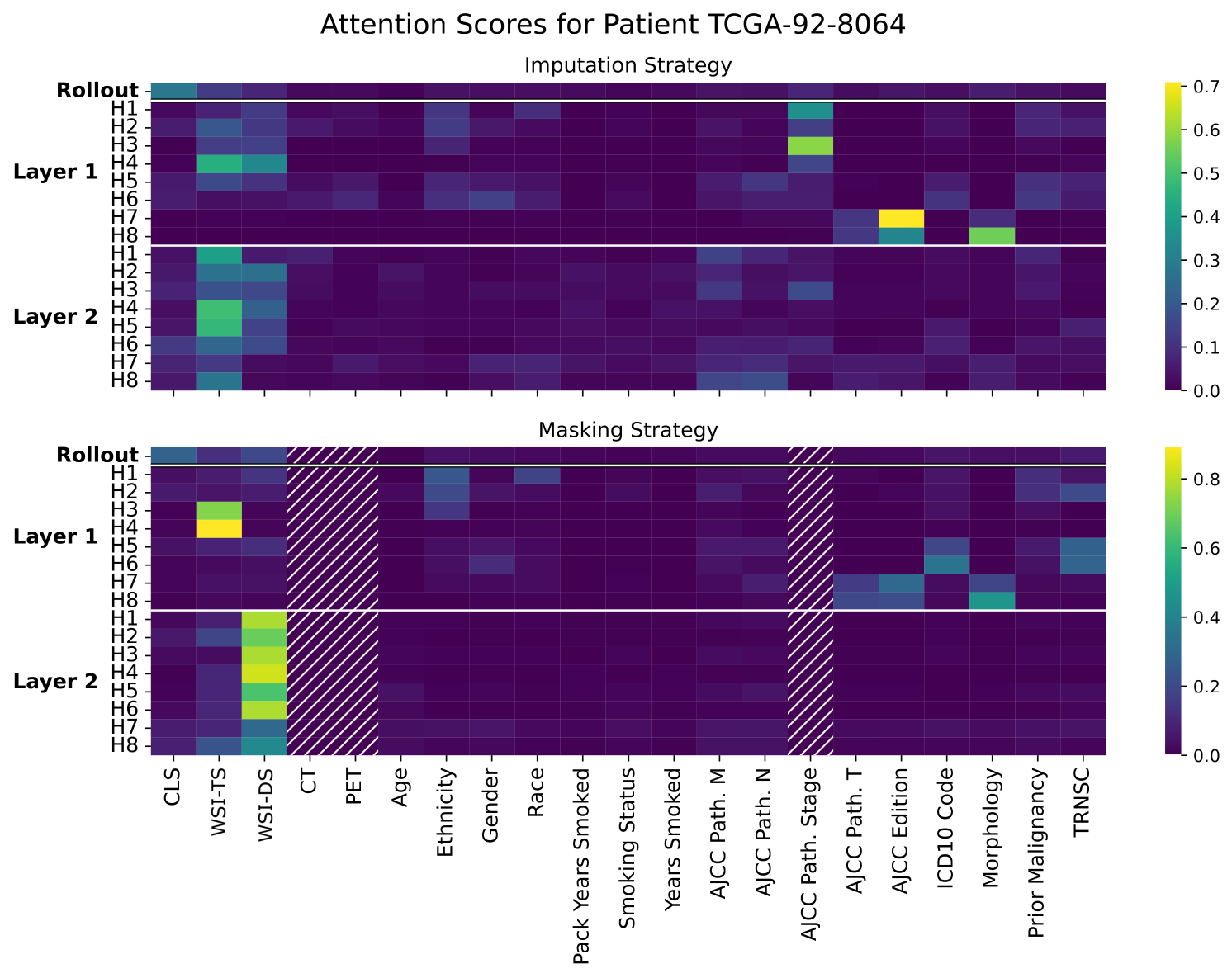}
    \caption{Attention scores for a Lung cancer patient comparing imputation (top) and masking (bottom) approaches. Missing entries are indicated by hatched cells, to emphasize the differences in attention assigned to these variable by the two approaches.}
    \label{Lung-TCGA-attn}
\end{figure}

We can further extend this analysis by comparing masking and imputation strategies. Figure~\ref{Lung-TCGA-attn} shows how the attention scores change depending on the strategy for a Lung cancer patient. Both models produce the correct prediction, but their attention patterns differ substantially. Despite the AJCC tumour stage being missing, the imputation model assigns it high attention, potentially indicating reliance on information introduced through imputation or on dataset-specific associations. In contrast, other missing variables, such as CT and PET data, receive little attention.

In contrast, the masking model restricts attention to observed information and places greater emphasis on the available WSI data. This suggests that masking may provide a more reliable representation of missingness, allowing the model to leverage complementary modalities when important clinical information is unavailable.

\subsection{Use of Temporal Knowledge for Binary Classification}
While the previous binary classification experiments for the Lung cancer cohort considered a baseline scenario in which only modalities available at diagnosis were used, the temporal structure of the Lung and Gastric cancer datasets allows us to incorporate subsequent treatment and follow-up events when predicting overall survival. Compared to the previous Lung cancer binary survival results in Table~\ref{tab:attn-imp}, incorporating this additional information into each patient's timeline improves performance. Regarding positional encodings, the different approaches produced very similar results, as can be seen in Table~\ref{temporal-binary-classification}. For this particular task, explicit knowledge of the absolute timing of events may be less informative than the content of the events themselves. Consequently, incorporating different forms of positional information does not lead to substantial performance differences.

\begin{table}[]
\centering
\caption{Longitudinal overall survival classification for Lung and Gastric cancer cohorts using two positional encoding approaches.}
\label{temporal-binary-classification}
\fontsize{8pt}{9.6pt}\selectfont
\begin{tabular}{@{}ccccc@{}}
\toprule
        & \multicolumn{2}{c}{{\color[HTML]{000000} Learnable PE}} & \multicolumn{2}{c}{Sinusoidal PE} \\ \cmidrule(l){2-5} 
\multirow{-2}{*}{\begin{tabular}[c]{@{}c@{}}Cancer\\ Cohort\end{tabular}} & BalAcc {[}\%{]} & ROCAUC {[}\%{]} & BalAcc {[}\%{]} & ROCAUC {[}\%{]} \\ \midrule
Lung    & {\color[HTML]{000000} 73.35 ± 4.61}    & 75.50 ± 6.06   & 75.23 ± 4.78    & 76.88 ± 5.57    \\
Gastric & 71.45 ± 4.59                           & 79.01 ± 5.52   & 70.42 ± 5.00    & 80.52 ± 5.97    \\ \bottomrule
\end{tabular}
\end{table}

\subsection{Discrete-Time Survival Prediction}

\begin{figure}
    \centering
    \includegraphics[width=0.9\linewidth]{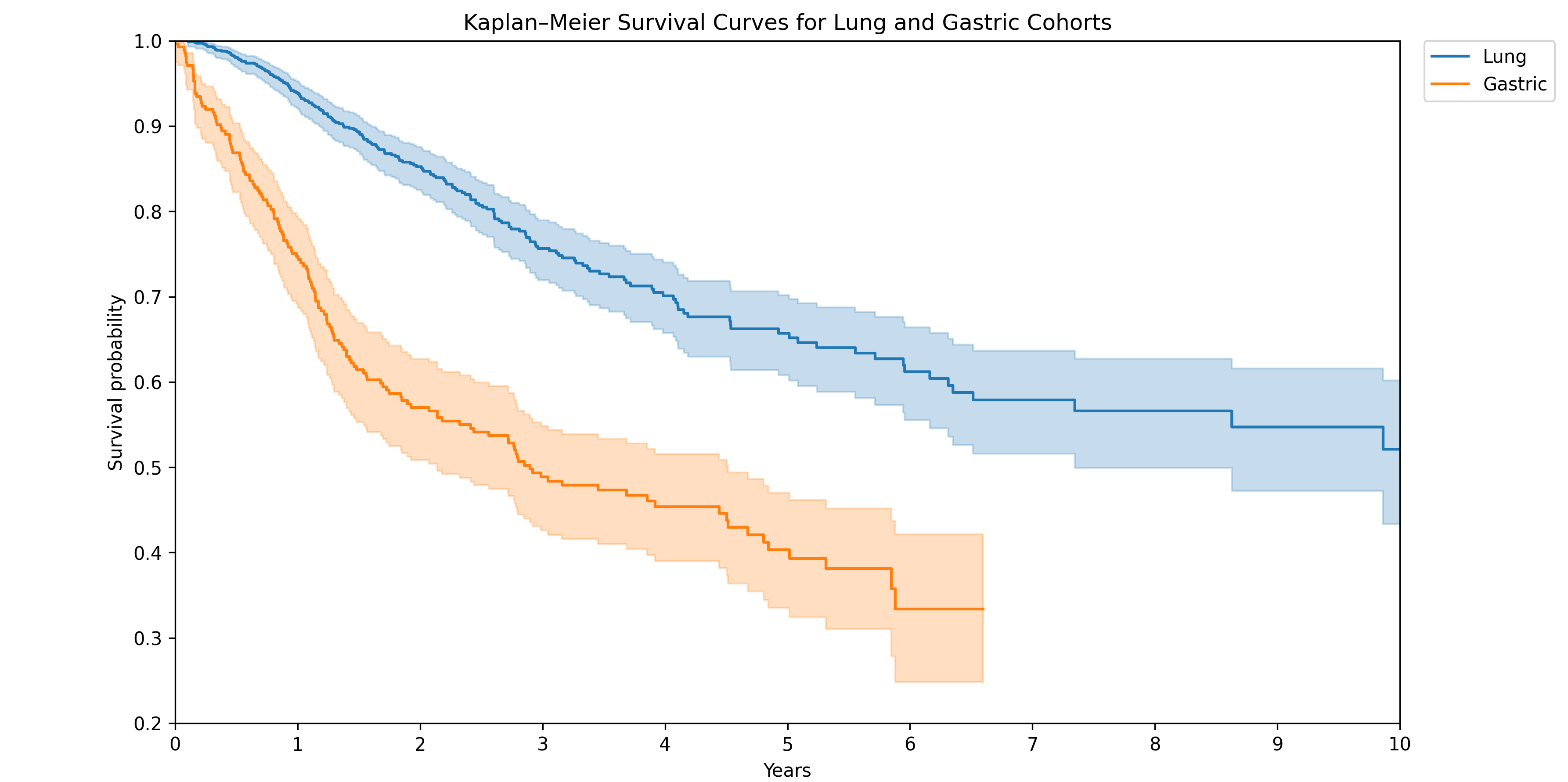}
    \caption{Kaplan-Meier survival curves for Lung and Gastric cancer cohorts.}
    \label{km-Lung-Gastric}
\end{figure}

For the longitudinal analysis, we use the Lung and Gastric cohorts. Figure~\ref{km-Lung-Gastric} shows their Kaplan-Meier survival curves, providing an overall, purely statistic, non-personalized view of the survival distributions. The Gastric cohort has a maximum follow-up of approximately 6 years, with estimated survival reaching 50\% at around the 3-year mark. For the Lung cancer cohort, 671 of the 1,365 patients have an available censoring or last follow-up date and are therefore included in the longitudinal analysis. Despite the relatively long follow-up, only 10\% of these patients experienced the event, resulting in a high estimated survival probability.

For discrete-time survival analysis, we compare predictions made at each patient's last known event with predictions from a common one-year landmark, using all information available up to the respective landmark. Results are shown in Table~\ref{concordance-index-DHP}. For Gastric, learnable and sinusoidal positional encodings produce similar performance. For Lung, learnable positional encoding performs better in the last-event setting, particularly in terms of C-index.

\begin{table}[]
\centering
\caption{Learnable and Sinusoidal positional encodings for predicting patient survival at last known patient event or at a fixed landmark at the 1-year mark.}
\label{concordance-index-DHP}
\fontsize{8pt}{9.6pt}\selectfont
\begin{tabular}{@{}cccccc@{}}
\toprule
        &                    & \multicolumn{2}{c}{{\color[HTML]{000000} Learnable PE}} & \multicolumn{2}{c}{Sinusoidal PE} \\ \cmidrule(l){3-6} 
\multirow{-2}{*}{\begin{tabular}[c]{@{}c@{}}Cancer\\ Cohort\end{tabular}} &
  \multirow{-2}{*}{Prediction} &
  C-Index {[}\%{]} &
  ROCAUC {[}\%{]} &
  C-Index {[}\%{]} &
  ROCAUC {[}\%{]} \\ \midrule
Lung    & Last Patient Event & {\color[HTML]{000000} 68.05 ± 5.59}    & 57.73 ± 5.78   & 66.84 ± 10.11    & 58.60 ± 5.10   \\
        & Fixed Landmark @1 Year            & 64.61 ± 1.52                           & 44.05 ± 2.54   & 64.14 ± 1.80     & 48.23 ± 3.18   \\ \midrule
Gastric & Last Patient Event & 85.53 ± 3.36                           & 88.51 ± 3.99   & 85.40 ± 2.83     & 88.30 ± 3.19   \\
        & Fixed Landmark @1 Year            & 75.25 ± 3.43                           & 79.45 ± 4.59   & 76.40 ± 2.38     & 80.95 ± 2.92   \\ \bottomrule
\end{tabular}
\end{table}

We next compare the Transformer models with a Kaplan-Meier (KM) baseline using Brier scores. For patient-specific landmarks, predictions are made at relative horizons of one to five years using only information available up to each patient's landmark. For the fixed-landmark setting, only patients who remain event-free at one year are included. Because positional encoding strategies show little difference in this setting, the subsequent results use learnable temporal encoding.

\begin{figure}
    \centering
    \includegraphics[width=0.6\linewidth]{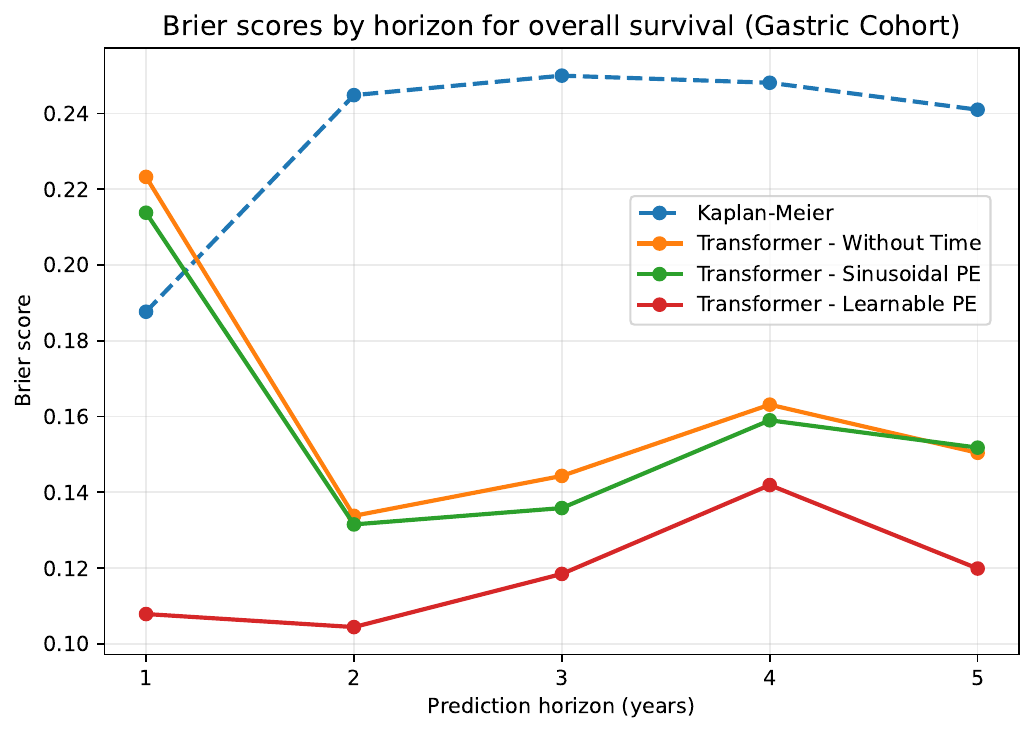}
    \caption{Brier score comparison using Kaplan-Meier and Transformer for a set of prediction landmarks.}
    \label{Gastric-overall-survival-brierscore}
\end{figure}

Figure~\ref{Gastric-overall-survival-brierscore} shows the Brier-score comparison for the Gastric cohort using patient-specific landmarks. Transformer models achieve lower Brier scores than the KM baseline across most prediction horizons, with the learnable temporal encoding generally performing best. This suggests that explicitly modelling elapsed time is beneficial when patient-specific landmarks are used.

To ensure a more comparable scenario, rather than making predictions at each patient’s last known event, we defined a common landmark at 1 year. Accordingly, only patients who remained event-free at the 1-year landmark were included in the analysis. Here however, because the predictions were all made at the same landmark, the use of different positional encodings had little impact on the results, hence, the results showed in the subsequent Tables are using learnable encoding.

Table~\ref{Lung-brier} presents the Brier score results for two patient cohorts from the Lung dataset: a high-risk cohort and patients that were alive at the one-year mark. Patients were stratified at diagnosis (0-day landmark) according to their AJCC TNM tumor staging, with 436 patients classified as high risk. When evaluated on this cohort, the Transformer-based model achieves lower Brier scores than KM from years 2 through 5. In contrast, performance is less consistent in the event-free cohort, where KM remains competitive. One possible explanation is the strong class imbalance in this cohort. As illustrated by the KM estimates, relatively few patients experience the event during the prediction period, resulting in a survival distribution dominated by non-events.

The stronger performance observed for the high-risk cohort suggests that the model benefits from a cohort in which clinically relevant risk factors are more pronounced and the survival outcomes provide greater differentiation between patients. In contrast, the event-free cohort at the 365-day landmark represents a more homogeneous group with relatively few subsequent events, making patient-level risk discrimination more challenging.

\begin{table}[]
\centering
\caption{Brier scores for the Lung cancer cohort, comparing Transformer-based models with the Kaplan-Meier (KM) baseline in a high-risk patient cohort at diagnosis according to AJCC tumor stage, and among patients who remained event-free at the 365-day landmark. \textbf{Bold} denotes timestamps that Transformer performed better.}
\label{Lung-brier}
\fontsize{8pt}{9.6pt}\selectfont
\begin{tabular}{@{}ccccccccc@{}}
\toprule
 &
  {\color[HTML]{000000} } &
   &
   &
  \multicolumn{5}{c}{{\color[HTML]{000000} Brier-Score}} \\ \cmidrule(l){5-9} 
\multirow{-2}{*}{} &
  \multirow{-2}{*}{{\color[HTML]{000000} Cohort}} &
  \multirow{-2}{*}{n} &
  \multirow{-2}{*}{Landmark} &
  L+1 &
  L+2 &
  L+3 &
  L+4 &
  L+5 \\ \midrule
Transformer &
  {\color[HTML]{000000} } &
   &
   &
  {\color[HTML]{000000} 0.225} &
  \textbf{0.080} &
  \textbf{0.121} &
  \textbf{0.124} &
  \textbf{0.110} \\
Kaplan-Meier &
  \multirow{-2}{*}{{\color[HTML]{000000} High Risk}} &
  \multirow{-2}{*}{436} &
  \multirow{-2}{*}{L = 0} &
  0.039 &
  0.088 &
  0.136 &
  0.156 &
  0.163 \\ \midrule
Transformer &
   &
   &
   &
  0.323 &
  0.152 &
  0.144 &
  0.125 &
  \textbf{0.120} \\
Kaplan-Meier &
  \multirow{-2}{*}{\begin{tabular}[c]{@{}c@{}}Event-free\\ @1-Year\end{tabular}} &
  \multirow{-2}{*}{534} &
  \multirow{-2}{*}{L = 365} &
  0.056 &
  0.098 &
  0.115 &
  0.121 &
  0.135 \\ \bottomrule
\end{tabular}
\end{table}

For Gastric, Table~\ref{Gastric-brier} shows that the Transformer outperforms the KM baseline at most prediction horizons for both the full cohort and the one-year event-free cohort. The larger differences relative to Lung may reflect the shorter follow-up and more balanced event distribution in the Gastric cohort.

\begin{table}[]
\centering
\caption{Brier scores for the Gastric cancer cohort, comparing Transformer-based models with the Kaplan-Meier baseline at different landmarks. \textbf{Bold} denotes timestamps that Transformer performed better.}
\label{Gastric-brier}
\fontsize{8pt}{9.6pt}\selectfont
\begin{tabular}{@{}ccccccccc@{}}
\toprule
{\color[HTML]{000000} }            &  &  & {\color[HTML]{000000} } & \multicolumn{5}{c}{Brier-Score}                                                    \\ \cmidrule(l){5-9} 
\multirow{-2}{*}{{\color[HTML]{000000} }} &
  \multirow{-2}{*}{Cohort} &
  \multirow{-2}{*}{n} &
  \multirow{-2}{*}{{\color[HTML]{000000} Landmark}} &
  L+1 &
  L+2 &
  L+3 &
  L+4 &
  L+5 \\ \midrule
{\color[HTML]{000000} Transformer} &  &  & {\color[HTML]{000000} } & 0.078          & \textbf{0.095} & \textbf{0.127} & \textbf{0.207} & 0.247          \\
Kaplan-Meier &
  \multirow{-2}{*}{\begin{tabular}[c]{@{}c@{}}All\\ Patients\end{tabular}} &
  \multirow{-2}{*}{274} &
  \multirow{-2}{*}{{\color[HTML]{000000} L = 0}} &
  0.070 &
  0.197 &
  0.253 &
  0.256 &
  0.233 \\ \midrule
Transformer                        &  &  &                         & \textbf{0.094} & \textbf{0.167} & \textbf{0.183} & \textbf{0.231} & \textbf{0.204} \\
Kaplan-Meier &
  \multirow{-2}{*}{\begin{tabular}[c]{@{}c@{}}Event-free\\ @1-Year\end{tabular}} &
  \multirow{-2}{*}{202} &
  \multirow{-2}{*}{L = 365} &
  0.099 &
  0.180 &
  0.249 &
  0.262 &
  0.213 \\ \bottomrule
\end{tabular}
\end{table}

\subsection{Discrete-Time Survival Interventions}

\begin{figure}[!htb]
    \centering
    % Top pair
    \begin{tabular}{cc}
        \includegraphics[width=.45\textwidth]{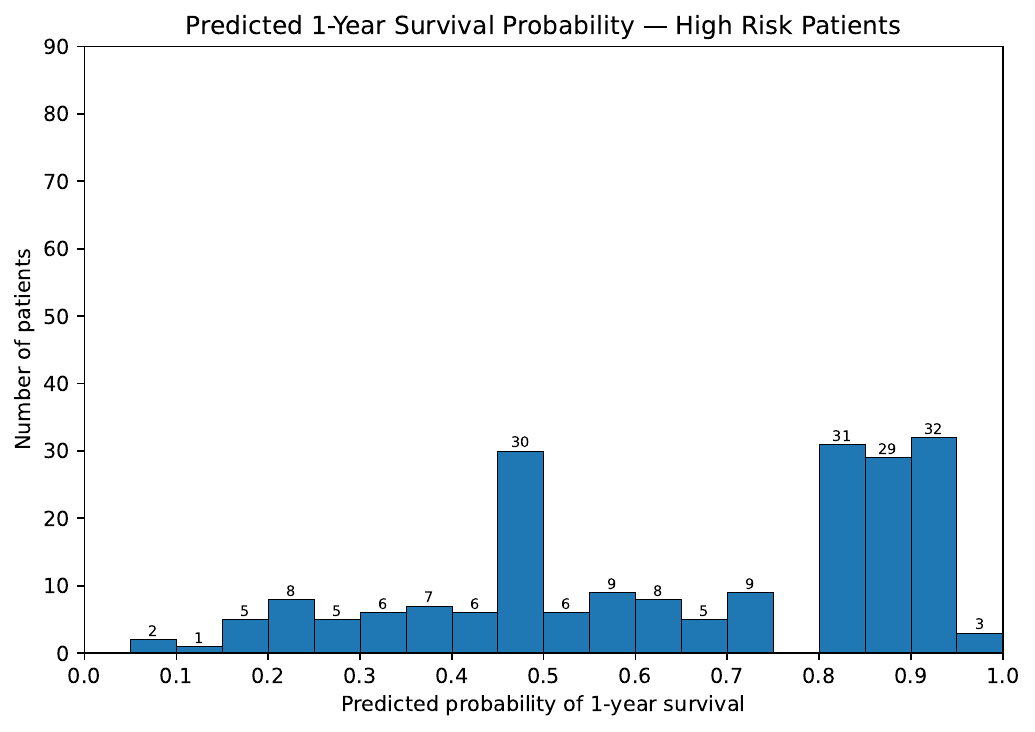} & \includegraphics[width=.45\textwidth]{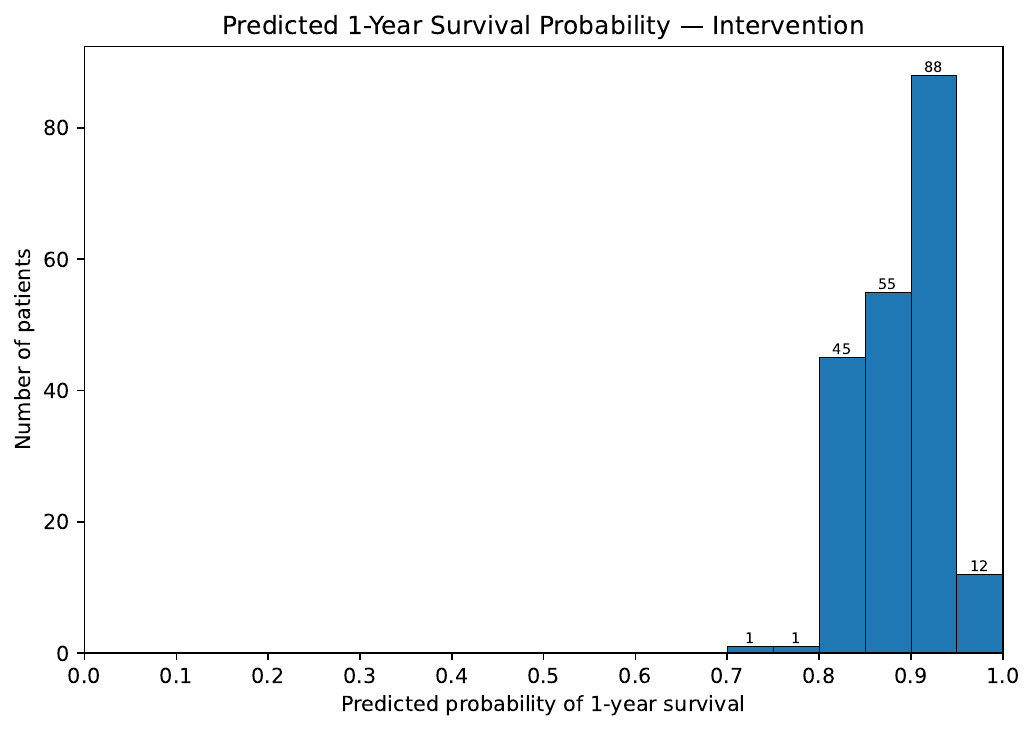}
    \end{tabular}
    \caption{Intervention results for the high-risk cohort after setting the AJCC tumor characteristics to their minimum values: (left) original predicted probabilities and (right) post-intervention predicted probabilities.}
        \label{fig:highrisk-intervention-results}
\end{figure}

\begin{figure}[!htb]
    \centering
    % Top pair
    \begin{tabular}{cc}
        \includegraphics[width=.45\textwidth]{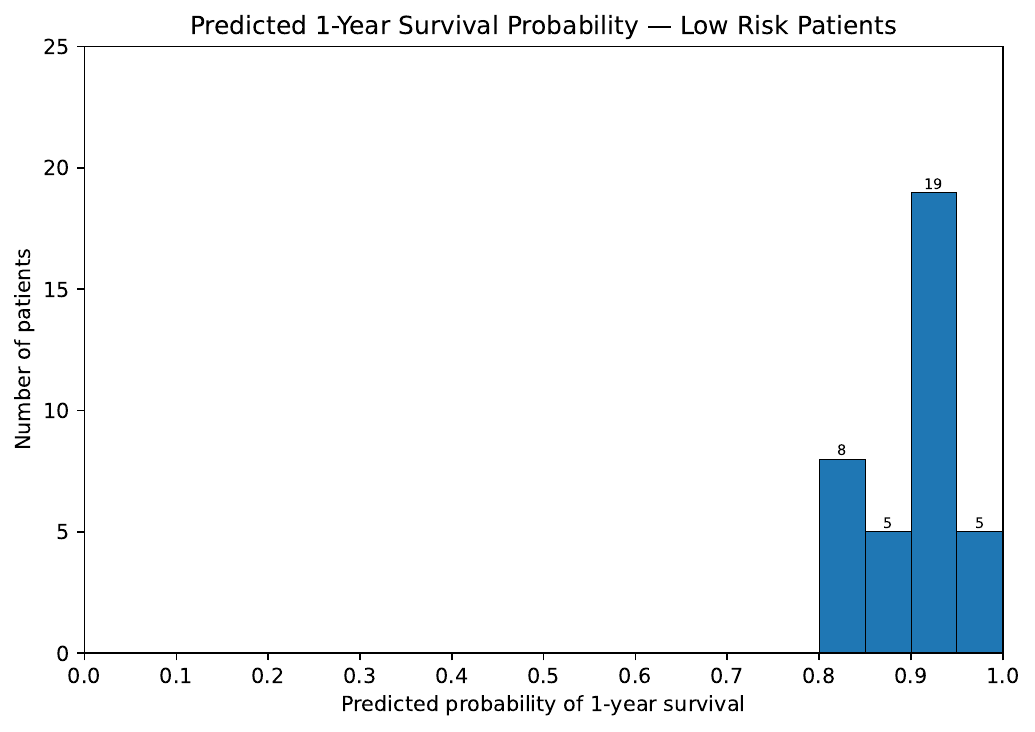} & \includegraphics[width=.45\textwidth]{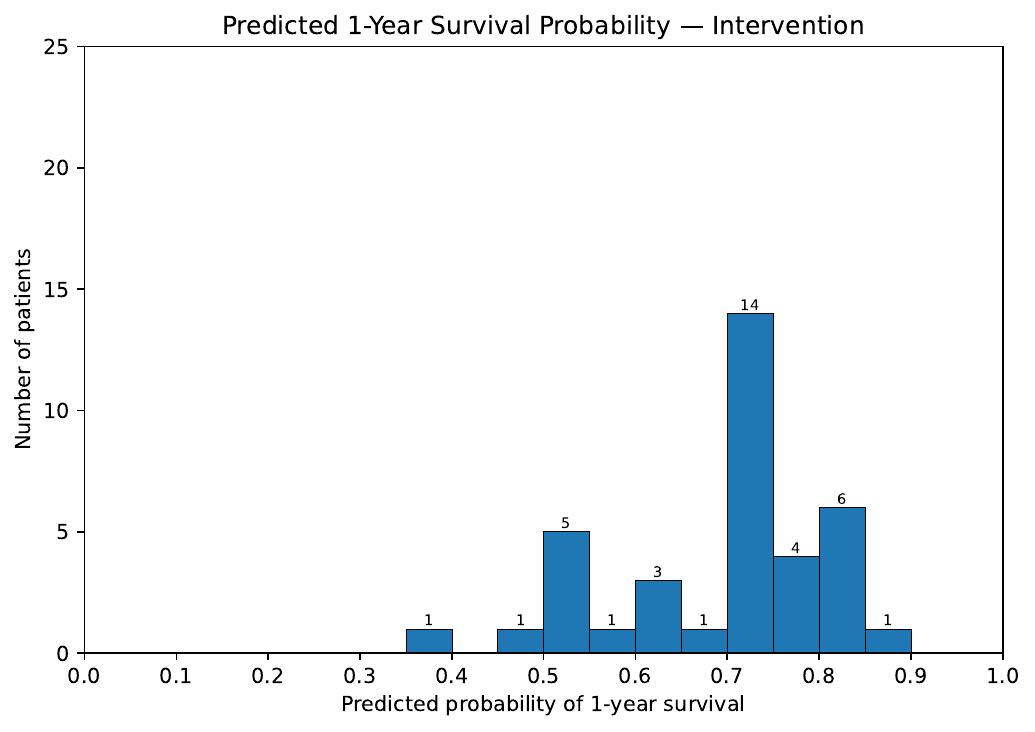}
    \end{tabular}
    \caption{Intervention results for the low-risk cohort after setting the AJCC tumor characteristics to their maximum values: (left) original predicted probabilities and (right) post-intervention predicted probabilities.}
        \label{fig:lowrisk-intervention-results}
\end{figure}

To further investigate the explainability of the discrete-time survival model, we performed a series of feature interventions on the Gastric cancer cohort. At diagnosis, patients were classified as high- or low-risk based on AJCC T, N, and M staging, resulting in 202 and 37 patients, respectively. A small subset could not be classified due to missing data. Figures~\ref{fig:highrisk-intervention-results} and \ref{fig:lowrisk-intervention-results} show the original and intervened 1-year survival predictions for both cohorts. For the high-risk cohort, AJCC T, N, and M stages were set to their minimum possible values, while for the low-risk cohort they were set to their maximum values, shifting each patient towards the opposite risk profile while keeping all other characteristics unchanged. The resulting changes in predicted survival demonstrate the model's sensitivity to clinically relevant staging variables and its ability to reflect the expected direction of changes in risk.


    %\caption{Intervention results for the high-risk (top) and low-risk (bottom) Gastric cancer patient cohort. On the left, the original predicted probabilities and on the right the post-intervention predicted probabilities.}
    %\label{fig:intervention-results}

\begin{figure}[!htb]
    \centering
    \includegraphics[width=0.8\linewidth]{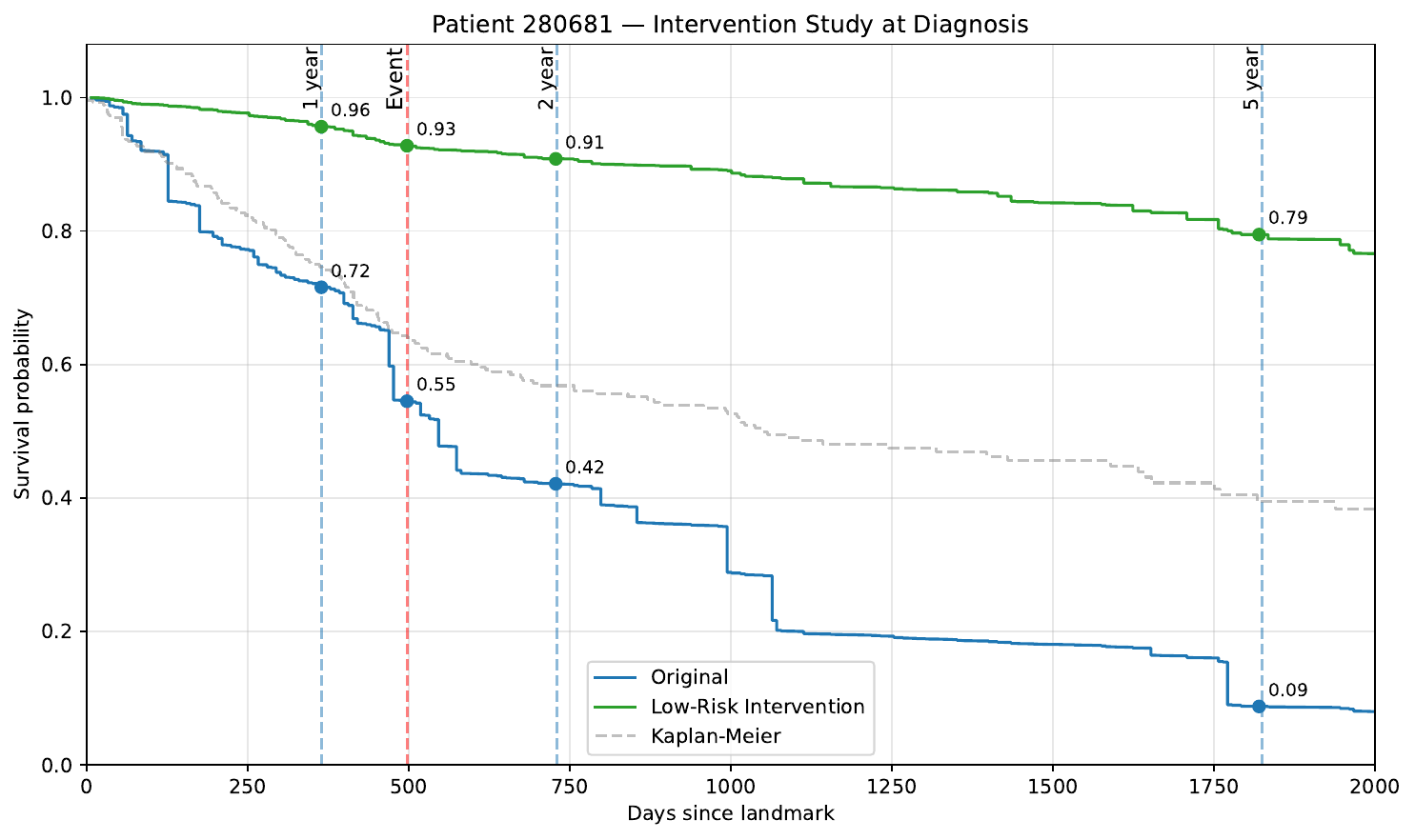}
    \caption{Survival curves for a high-risk Gastric cancer patient under the original tumor status and following an intervention in which the tumor status is changed to the best-case scenario.}
    \label{single-patient-study}
\end{figure}

While the previous analysis considered patients collectively, Figure~\ref{single-patient-study} illustrates the interventions for a single high-risk Gastric cancer patient. The patient presented with metastasis, the highest AJCC T stage, and positive N staging. Removing metastasis and setting the AJCC staging variables to their minimum values substantially shifts the predicted 5-year survival curve, demonstrating the influence of these clinical characteristics. The Kaplan-Meier curve for the high-risk cohort is included for comparison with the original prediction. Despite the patient's high-risk classification, predicted survival remains above 60\% at the observed event time and falls below 50\% only after approximately three years of follow-up.

\section{Conclusion}

This work investigated Transformer-based models for fusing heterogeneous multimodal data and integrating longitudinal clinical observations, with a focus on tokenization, missing data, temporal information, and interpretability.

The results demonstrate the importance of explicitly accounting for missing information in multimodal clinical data. Discarding incomplete samples can lead to substantial data loss, while imputation may introduce noise. Feature-level tokenization enables missing features to be selectively masked and, when feasible, outperformed the alternative strategies considered, providing greater robustness to missing information. We further found substantial variation in attention scores across heads and layers, highlighting the limitations of interpreting individual attention maps. Attention rollout therefore provides a more suitable aggregated view for model explainability. Overall, the proposed Transformer-based approach effectively fused heterogeneous modalities and outperformed the evaluated unimodal baselines.

For longitudinal data, we compared conventional sinusoidal positional encoding with a learnable temporal encoding based on elapsed time. The improved performance of the latter suggests that explicitly representing temporal distances can be beneficial for longitudinal survival prediction. These findings highlight the potential of Transformer-based architectures for integrating heterogeneous and longitudinal clinical data while accounting for missing information and temporal structure.

%Future work could investigate whether these findings generalize to other Transformer architectures and datasets, with the aim of establishing broader design guidelines for multimodal learning under missing data.

% Use the style of numbering in square brackets.
% If nothing is used, default style will be taken.
%\begin{enumerate}[a)]
%\item 
%\item 
%\item 
%\end{enumerate}  

%\begin{itemize}
%\item 
%\item 
%\item 
%\end{itemize}  

%\begin{description}
%\item[]
%\item[] 
%\item[] 
%\end{description}  

%\clearpage %%Remove this from your manuscript

% Figure
%\begin{figure}%[]
%  \centering
%    \includegraphics{}
%    \caption{}\label{fig1}
%\end{figure}

%\begin{table}%[]
%\caption{}\label{tbl1}
%\begin{tabular*}{\tblwidth}{@{}LL@{}}
%\toprule
%  &  \\ % Table header row
%\midrule
% & \\
% & \\
% & \\
% & \\
%\bottomrule
%\end{tabular*}
%\end{table}

% Uncomment and use as the case may be
%\begin{theorem} 
%\end{theorem}

% Uncomment and use as the case may be
%\begin{lemma} 
%\end{lemma}


%\section{}\label{}


\section*{Acknowledgements}
This work was supported by LARSyS FCT funding (DOI: \nolinkurl{10.54499/LA/P/0083/2020} and \nolinkurl{10.54499/UID/50009/2025}), and the EU project NextGen.

\bibliographystyle{elsarticle-num} 
\bibliography{cas-refs}




% %% For numbered reference style

%   \textit{\LaTeX: a document preparation system},
%   Addison Wesley, Massachusetts,

% \end{thebibliography}
\end{document}